\documentclass[]{interact}

\usepackage{epstopdf}
\usepackage{subfigure}
\usepackage{tabularx}
\usepackage{graphicx}
\usepackage{amsmath}
\usepackage{amssymb}
\usepackage{booktabs}
\usepackage{placeins} 
\usepackage{float} 
\usepackage{enumitem}
\usepackage{tabularx} 
\usepackage{longtable} 
\usepackage{array} 
\usepackage{ragged2e} 
\usepackage{booktabs} 
\usepackage{placeins}
\usepackage{afterpage} 
\usepackage{caption} 
\usepackage{booktabs} 
\usepackage{adjustbox} 
\usepackage{graphicx} 
\usepackage{tabularx}

\theoremstyle{plain}

\theoremstyle{definition}

\theoremstyle{remark}

\begin{document}


\title{Efficient Dynamic Attention 3D Convolution for Hyperspectral Image Classification}

\author{
\name{Guandong Li\textsuperscript{a}\thanks{CONTACT Guandong Li. Email: leeguandon@gmail.com} and Mengxia Ye\textsuperscript{b}}
\affil{\textsuperscript{a}iFLYTEK, Shushan, Heifei, Anhui, China; \textsuperscript{b}Aegon THTF,Qinghuai,Nanjing,Jiangsu,China}
}

\maketitle

\begin{abstract}
Deep neural networks face several challenges in hyperspectral image classification, including insufficient utilization of joint spatial-spectral information, gradient vanishing with increasing depth, and overfitting. To enhance feature extraction efficiency while skipping redundant information, this paper proposes a dynamic attention convolution design based on an improved 3D-DenseNet model. The design employs multiple parallel convolutional kernels instead of a single kernel and assigns dynamic attention weights to these parallel convolutions. This dynamic attention mechanism achieves adaptive feature response based on spatial characteristics in the spatial dimension of hyperspectral images, focusing more on key spatial structures. In the spectral dimension, it enables dynamic discrimination of different bands, alleviating information redundancy and computational complexity caused by high spectral dimensionality. The DAC module enhances model representation capability by attention-based aggregation of multiple convolutional kernels without increasing network depth or width. The proposed method demonstrates superior performance in both inference speed and accuracy, outperforming mainstream hyperspectral image classification methods on the IN, UP, and KSC datasets.

\end{abstract}


\section{Introduction}

Hyperspectral remote sensing images play a crucial role in spatial information applications due to their unique narrow-band imaging characteristics. These image data typically contain dozens to hundreds of continuous spectral bands, enabling the acquisition of fine spectral features of target objects and significantly improving the accuracy of ground object recognition \cite{chang2003hyperspectral}. The imaging equipment not only records the spectral characteristics of each sampling point but also synchronously acquires its spatial position information, achieving organic integration of spatial geometric data and spectral features, ultimately constructing a three-dimensional data structure containing two-dimensional space and one-dimensional spectrum. As an important application direction of remote sensing technology, ground object recognition and classification have significant value in various fields, including ecological environment assessment, transportation planning, agricultural production monitoring, land resource management, and engineering geological surveys \cite{bing2011intelligent}.

In recent years, deep learning has been introduced for hyperspectral image (HSI) classification in remote sensing \cite{sun2016random, wang2019domain, gong2019cnn, safari2020multiscale}, achieving substantial progress. In \cite{lee2017going} and \cite{zhao2016spectral}, Principal Component Analysis (PCA) was first applied to reduce the dimensionality of the entire hyperspectral data, followed by extracting spatial information from neighboring regions of the input hyperspectral data using 2D CNN for feature extraction. \cite{yue2015spectral} employed deep CNN to extract deep features combined with logistic regression classifiers for classification. However, methods like 2D-CNN \cite{makantasis2015deep, chen2016deep} require separate extraction of spatial and spectral features, failing to fully utilize joint spatial-spectral information and necessitating complex preprocessing. \cite{wang2018fast} proposed a Fast Dense Spectral-Spatial Convolutional Network (FDSSC) based on dense networks, constructing 1D-CNN and 3D-CNN dense blocks connected in series to form a deep network. \cite{li2019doubleconvpool} directly used stacked 3D-CNN for classification based on a dual convolution-pooling structure, achieving promising results. FSKNet \cite{li2022faster} introduced a 3D-to-2D module and selective kernel mechanism, while 3D-SE-DenseNet \cite{li2020hyperspectral} incorporated the SE mechanism into 3D-CNN to correlate feature maps between different channels, activating effective information while suppressing ineffective information in feature maps. DGCNet \cite{li2023dgcnet} designed a dynamic grouped convolution (DGC) on 3D convolution kernels, where DGC introduces small feature selectors for each group to dynamically determine which part of input channels to connect based on activations of all input channels. Multiple groups can capture different complementary visual/semantic features of input images, enabling CNNs to learn rich feature representations. Therefore, using 3D-CNN to extract spatial and spectral information from hyperspectral remote sensing images has become an important direction for hyperspectral remote sensing image classification. DHSNet \cite{liu2025dual} proposed a novel Central Feature Attention-Aware Convolution (CFAAC) module that guides attention to focus on central features crucial for capturing cross-scene invariant information. By concentrating on these key features, the model's ability to represent generalized features is enhanced, making it suitable for cross-scene classification tasks. LGCNet \cite{li2025spatial} improved the drawbacks of grouped convolution by incorporating ideas from dynamic learning networks, designing learnable grouped convolution on 3D convolution kernels where both input channels and convolution kernel groups can be learned, allowing flexible grouping structures and producing better representation capabilities. To leverage the advantages of both CNN and Transformer, many studies have attempted to combine CNN and Transformer to utilize local and global feature information of HSI. \cite{sun2022spectral} proposed a Spectral-Spatial Feature Tokenization Transformer (SSFTT) network that extracts shallow features through 3D and 2D convolutional layers and uses Gaussian-weighted feature tokens to extract high-level semantic features in the transformer encoder. Some Transformer-based methods \cite{hong2021spectralformer} employ grouped spectral embedding and transformer encoder modules to model spectral representations, but these methods have obvious shortcomings. These improved Transformer methods treat spectral bands or spatial patches as tokens and encode all tokens, resulting in significant redundant computations. However, HSI data already contains substantial redundant information, and their accuracy often falls short compared to 3D-CNN-based methods while requiring greater computational complexity than 3D-CNN.

3D-CNN possesses the capability to sample simultaneously in both spatial and spectral dimensions, maintaining the spatial feature extraction ability of 2D convolution while significantly reducing the parameter scale of convolutional kernels and ensuring effective spectral feature extraction. 3D-CNN can directly process high-dimensional data, eliminating the need for preliminary dimensionality reduction of hyperspectral images. However, introducing the spectral dimension into convolutional kernels dramatically increases the number of parameters in feature maps, especially when the input spectral dimension is high, leading to exponential growth in computational complexity and model parameters. Reducing the use of 3D convolution typically impairs the network's feature extraction capability in both spatial and spectral dimensions, resulting in substantial computational requirements. Currently widely used methods such as DFAN \cite{zhang2020deep}, MSDN \cite{zhang2019multi}, 3D-DenseNet \cite{zhang2019three}, and 3D-SE-DenseNet employ operations like dense connections. Dense connections directly link each layer to all its preceding layers, enabling feature reuse, but they introduce redundancy when subsequent layers do not require early features. Therefore, how to more efficiently enhance the representational capability of 3D convolution kernels in 3D convolution, achieving more effective feature extraction with fewer cascaded 3D convolution kernels and dense connections, has become a direction in hyperspectral classification.

This paper proposes a novel operator design called dynamic convolution to improve representation capability with negligible additional floating-point operations (FLOPs). Dynamic convolution uses a set of K parallel convolutional kernels $\{\tilde{W}_k, \tilde{b}_k\}$ instead of a single convolutional kernel per layer. These convolutional kernels are dynamically aggregated through an input-dependent attention mechanism $\pi_k(x)$: $\tilde{W} = \sum_k \pi_k(x)\tilde{W}_k$ for each individual input feature. The bias terms are aggregated using the same attention mechanism: $\tilde{b} = \sum_k \pi_k(x)\tilde{b}_k$. Dynamic convolution is a nonlinear function with stronger representational capability than its static counterpart. Simultaneously, dynamic convolution is computationally efficient. It does not increase network depth or width because parallel convolutional kernels share output channels through aggregation. It only introduces additional computational costs for calculating attention $\{\pi_k(x)\}$ and aggregating kernels, which are negligible compared to convolution. The key point is that, at reasonable model size costs (since convolutional kernels are small), dynamic kernel aggregation provides an efficient way (with low additional FLOPs) to enhance representational capability. This paper introduces a dynamic attention mechanism in 3D convolutional kernels. First, in the spatial dimension, hyperspectral images' spatial features typically exhibit strong local correlations, such as texture, edge, and shape information of ground objects. The Dynamic Attention Convolution (DAC) introduces multiple parallel convolutional kernels $\{\tilde{W}_k\}$ and uses an attention mechanism $\pi_k(x)$ to dynamically aggregate these kernels, enabling adaptive adjustment of kernel weights based on input data's spatial characteristics. This dynamic nature allows the model to focus more flexibly on key spatial structures when processing different spatial regions, rather than relying on the fixed receptive field of a single static convolutional kernel. Compared to traditional 3D-CNN, this method enhances spatial feature representation capability through attention mechanisms without increasing network depth or width, especially when targets in hyperspectral images are unevenly distributed or have complex spatial patterns, enabling more efficient extraction of local features. Second, in the spectral dimension, the unique advantage of hyperspectral data lies in its rich band information, where each band reflects the target's reflection characteristics in a specific spectral range. This high-dimensional spectral information provides fine-grained discrimination basis for classification tasks. The DAC module can dynamically adjust the attention degree of convolutional kernels to different bands through the attention mechanism in the spectral dimension. It avoids the uniform sampling and dimensionality reduction treatment of all spectral dimensions in traditional 3D-CNN, effectively alleviating parameter redundancy and computational complexity caused by excessive spectral dimensions. The DAC module enables 3D-CNN to achieve sufficient feature extraction while balancing speed and computational requirements, and it is compatible with existing 3D-CNNs, achieving higher flexibility.

The main contributions of this paper are as follows:

1. This paper proposes an improved efficient 3D-DenseNet-based spatial-spectral joint hyperspectral image classification method using the DAC module. It employs 3D-CNN as the basic structure combined with lightweight and efficient dynamic convolution design, integrating dense connections with the DAC module. Dense connections facilitate feature reuse in the network. The designed 3D-DenseNet model combined with DAC achieves good accuracy on both IN and UP datasets with fewer parameters.

2. This paper introduces a dynamic attention mechanism in 3D-CNN, using multiple parallel convolutional kernels instead of a single kernel and assigning dynamic attention weights to these parallel convolutions. This dynamic attention mechanism achieves dynamic response in the spatial dimension of hyperspectral images and dynamic discrimination of different bands in the spectral dimension, effectively skipping the redundancy mechanism of 3D-CNN and achieving more efficient feature extraction capability.

3. DACNet is more concise than networks combining various DL mechanisms, without complex connections and concatenations, and requires less computation. Without increasing network depth or width, it improves model representation capability through attention-based aggregation of multiple convolutional kernels.

\section{Dynamic Attention Convolution-Based Classification Method}

\subsection{Efficient CNNs}

Efficient convolution design has always been at the core of model algorithms and inference fields. MobileNetV1 \cite{howard2017mobilenets} significantly reduced FLOPs by decomposing 3×3 convolution into depthwise convolution and pointwise convolution. MobileNetV2 \cite{sandler2018mobilenetv2} introduced inverted residuals and linear bottlenecks. MobileNetV3 \cite{howard2019searching} applied squeeze-and-excitation \cite{hu2018squeeze} in residual layers and adopted a platform-aware neural architecture approach \cite{tan2019mnasnet} to find the optimal network structure. ShuffleNet further reduced the computational complexity of 1×1 convolution through channel shuffle operations. ShiftNet \cite{wu2018shift} replaced expensive spatial convolution with shift operations and pointwise convolution. In dynamic convolution, SkipNet \cite{wang2018skipnet} learned an additional controller using reinforcement learning to decide whether to skip. Compared to existing methods, our dynamic convolution can replace any static convolution kernel (e.g., 1×1, 3×3, depthwise separable convolution, grouped convolution) and is compatible with most CNN architectures. We use dynamic convolution kernels but the network structure is static, without employing complex search structures like NAS, and does not require additional controllers. The attention mechanism is embedded in each layer, enabling end-to-end training.

\subsection{Dynamic Attention 3D Convolution}

Hyperspectral image sample data is scarce and exhibits sparse ground object characteristics, with uneven spatial distribution and substantial redundant information in the spectral dimension. Although 3D-CNN structures can utilize joint spatial-spectral information, how to more effectively achieve deep extraction of spatial-spectral information remains a noteworthy issue. As the core of convolutional neural networks, convolution kernels are generally regarded as information aggregators that combine spatial information and feature dimension information in local receptive fields. Convolutional neural networks consist of a series of convolutional layers, nonlinear layers, and downsampling layers, enabling them to capture image features from a global receptive field for image description. However, training a high-performance network is challenging, and much work has been done to improve network performance from the spatial dimension perspective. For example, the Inception structure incorporates multi-scale information, aggregating features from different receptive fields to obtain performance gains. The Residual structure achieves deep network extraction by fusing features produced by different blocks. However, training such networks is costly, with slow convergence, and they are prone to overfitting on small-sample datasets. 3D-CNN also contains numerous redundant weights in feature extraction convolutions, with many weights contributing little to the final output. This paper introduces lightweight dynamic attention-based dynamic convolution in hyperspectral image processing, incorporating the DAC module into an improved efficient 3D-DenseNet. The DAC structure is shown in Figure 1. Dynamic convolution does not use a single convolution kernel per layer but dynamically aggregates multiple parallel convolution kernels based on their attention. Assembling multiple kernels is not only computationally efficient but also has stronger representational capability because these kernels are aggregated nonlinearly through attention. In practice, Global Average Pooling is first performed to obtain global spatial features, which are then mapped to K dimensions through two FC layers, followed by softmax normalization. The resulting K attention weights can then be assigned to the K kernels of that layer, treating the entire kernel as an attention object.

In DAC, our dynamic convolution cleverly uses the SE structure to generate an attention mechanism and employs attention weights to weight K convolution kernels. This achieves dynamic filtering in both spatial and spectral dimensions of hyperspectral images, focusing more on responses to key spatial structures in the spatial dimension and determining and skipping redundant spectral information in the spectral dimension. This is particularly effective for sparse ground objects in hyperspectral images, where spectral data is relatively scarce and belongs to the small-sample category. Feature maps obtained by different convolution kernels exhibit significant differences. Through DAC, combined with the depth characteristics of 3D-DenseNet, features can be extracted more effectively.

\subsubsection{Dynamic Perceptron}
The traditional perceptron is expressed as $y = g(W^T x + b)$, where $W$ is the weight matrix, $b$ is the bias vector, and $g$ is the activation function . We define a dynamic perceptron by aggregating multiple K linear functions $\{\tilde{W}_k^T x + \tilde{b}_k\}$ as follows:

\begin{equation}
y = g(\tilde{W}^T(x)x + \tilde{b}(x))
\end{equation}
where
\begin{equation}
\tilde{W}(x) = \sum_{k=1}^K \pi_k(x)\tilde{W}_k, \quad \tilde{b}(x) = \sum_{k=1}^K \pi_k(x)\tilde{b}_k
\end{equation}
\begin{equation}
0 \leq \pi_k(x) \leq 1, \quad \sum_{k=1}^K \pi_k(x) = 1
\end{equation}

Here, $\pi_k$ is the attention weight for the k-th linear function $\tilde{W}_k^T x + \tilde{b}_k$. Note that the aggregation weights $\tilde{W}(x)$ and bias $\tilde{b}(x)$ are functions of the input and share the same attention.

The attention weights $\{\pi_k(x)\}$ are not fixed but vary for each input x. They represent the optimal aggregation of linear models $\{\tilde{W}_k^T x + \tilde{b}_k\}$ given the input. The aggregated model $\tilde{W}^T(x)x + \tilde{b}(x)$ is a nonlinear function. Therefore, the dynamic perceptron has greater representational capability than its static counterpart.

Compared to the static perceptron, the dynamic perceptron has the same number of output channels but a larger model size. It also introduces two additional computations: (a) calculating attention weights $\{\pi_k(x)\}$; (b) aggregating parameters based on attention $\sum_k \pi_k \tilde{W}_k$ and $\sum_k \pi_k \tilde{b}_k$. The additional computational cost should be significantly lower than the cost of computing $\tilde{W}^T x + \tilde{b}$. Mathematically, the computational constraints can be expressed as:

\begin{equation}
O(\tilde{W}^T x + \tilde{b}) \gg O(\sum \pi_k \tilde{W}_k) + O(\sum \pi_k \tilde{b}_k) + O(\pi(x))
\end{equation}

where $O(\cdot)$ represents computational cost.

\begin{figure}[h]
\centering
\includegraphics[width=0.8\linewidth]{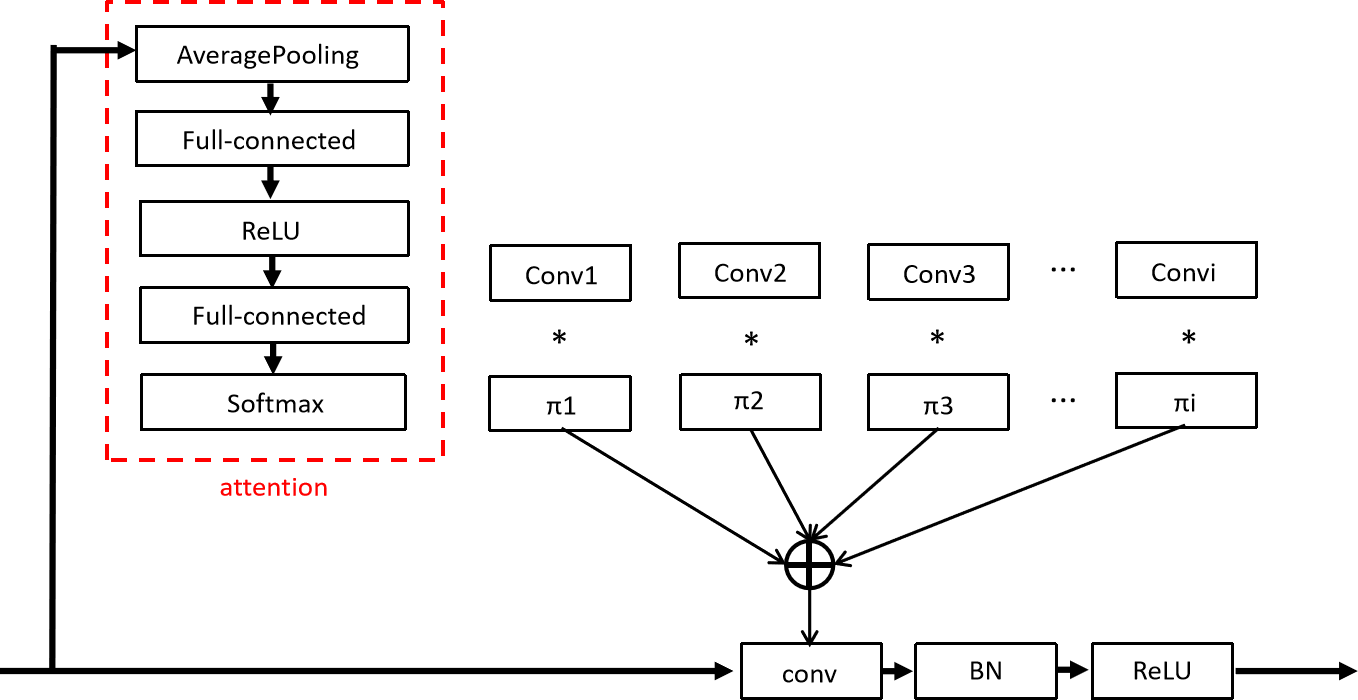}
\caption{The DAC module structure}
\label{fig:dac_module}
\end{figure}

\subsubsection{Dynamic Convolution}
We demonstrate a specific dynamic perceptron that satisfies the computational constraints - dynamic convolution. Similar to the dynamic perceptron, dynamic convolution employs $K$ convolutional kernels sharing identical kernel sizes and input/output dimensions. These kernels are aggregated using attention weights $\{\pi_k\}$. Following the aggregated convolution, we apply batch normalization and activation functions to construct dynamic convolutional layers.

We employ a Squeeze-and-Excitation (SE) inspired structure to compute the kernel attention $\{\pi_k(x)\}$. The global spatial information is first compressed through global average pooling. We then utilize two fully connected layers (with ReLU activation in between) followed by softmax to generate normalized attention weights for the $K$ convolutional kernels, where the first fully connected layer reduces the dimensionality by a factor of 4. 

Unlike SENet which computes attention at the channel level, our approach applies attention mechanism to the entire convolutional kernel ($\pi_k$). The computational cost of attention $O(\pi(x)) = HWC_{in} + C_{in}^2/4 + C_{in}K/4$ is significantly lower than the convolution's computational cost $O(\tilde{W}^Tx+\tilde{b}) = HWC_{in}C_{out}D_k^2$, where $D_k$ is the kernel size and $C_{out}$ is the number of output channels.

Aggregating $K$ convolutional kernels of size $D_k\times D_k$ with $C_{in}$ input channels and $C_{out}$ output channels introduces additional computational overhead of $KC_{in}C_{out}D_k^2 + KC_{out}$ multiply-add operations. Compared to the convolution's computational cost $HWC_{in}C_{out}D_k^2$, this additional cost becomes negligible when $K \ll HW$. 

The key innovation transforms a fixed convolutional kernel into an adaptive one determined by $K$ sub-kernels. The implementation pipeline consists of: (1) performing average pooling to obtain global spatial features, (2) mapping them to $K$ dimensions through two FC layers, and (3) applying softmax normalization to generate $K$ attention weights for kernel aggregation. This approach differs fundamentally from SENet's channel-wise attention by treating each complete kernel as an attention unit.

\begin{figure}[h]
\centering
\includegraphics[width=0.8\linewidth]{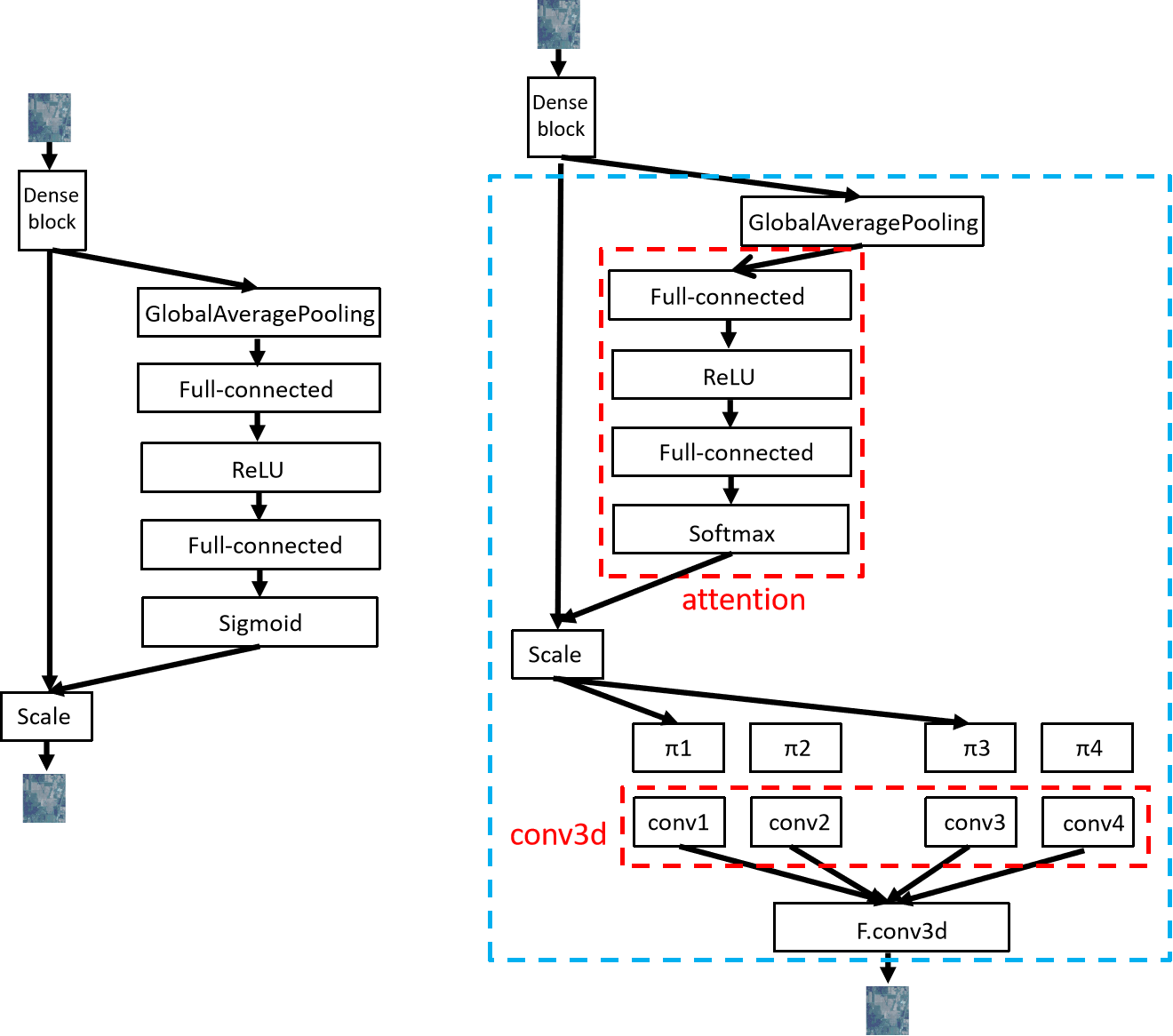}
\caption{Comparison between SE module (left) and DAC module (right)}
\label{fig:dac_structure}
\end{figure}

In our hyperspectral network architecture, the correspondence with SE modules is illustrated in Figure~\ref{fig:dac_structure}. As shown in the right panel of Figure~\ref{fig:dac_structure}, our dynamic convolution ingeniously adapts the SE structure to generate attention mechanisms, utilizing attention weights to perform weighted aggregation of $K$ convolutional kernels. This approach achieves dynamic filtering in both spatial and spectral dimensions of hyperspectral data, enhancing focus on critical spatial structures while enabling intelligent identification and skipping of redundant spectral information.

\subsection{3D-CNN Framework for Hyperspectral Image Feature Extraction}

\subsubsection{Training Network}

During DAC network training, the initial learning rate is set to 0.1 and decreases by a factor of 10 at the 30th, 60th, and 90th iterations. The weight decay is 1e-4. All models are trained for 100 epochs using the SGD optimizer with a momentum of 0.9. A dropout rate of 0.1 is applied before the final layer of DACNet.

\subsubsection{DGC-DenseNet Architecture Design}
We introduce two key modifications to the original 3D-DenseNet architecture to enhance simplicity and computational efficiency:

\textbf{Exponentially increasing growth rate.} The original DenseNet design adds $k$ new feature maps per layer, where $k$ is a constant growth rate. As shown in \cite{huang2017densely}, deeper layers in DenseNet tend to rely more on high-level features than low-level ones, which motivated our improvement through strengthened short connections. We achieve this by progressively increasing the growth rate with depth, enhancing the proportion of features from later layers relative to earlier ones. For simplicity, we set the growth rate as $k=2^{m-1}k_0$, where $m$ is the dense block index and $k_0$ is a constant. This growth rate configuration introduces no additional hyperparameters. The "increasing growth rate" strategy places a larger proportion of parameters in the model's later layers, significantly improving computational efficiency while potentially reducing parameter efficiency in some cases. Depending on specific hardware constraints, trading one for the other may be advantageous \cite{liu2017learning}.

\begin{figure}[h]
\centering
\includegraphics[width=0.9\linewidth]{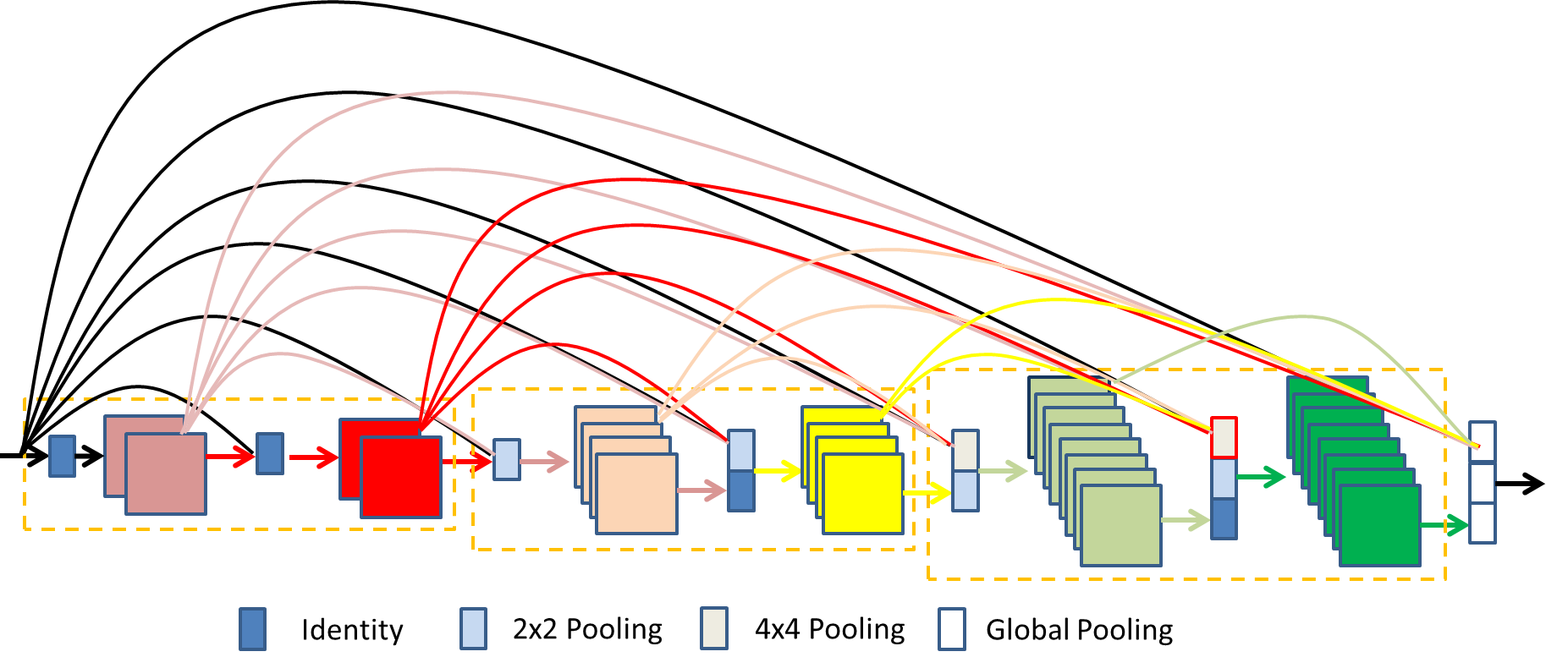}
\caption{Proposed DenseNet variant with two key differences from original DenseNet: (1) Direct connections between layers with different feature resolutions; (2) Growth rate doubles when feature map size reduces (third yellow dense block generates significantly more features than the first block)}
\label{fig:densenet_variant}
\end{figure}

\textbf{Fully dense connectivity.} To encourage greater feature reuse than the original DenseNet architecture, we connect the input layer to all subsequent layers across different dense blocks (see Figure~\ref{fig:densenet_variant}). Since dense blocks have different feature resolutions, we downsample higher-resolution feature maps using average pooling when connecting them to lower-resolution layers.

\begin{figure}[t]
\centering
\includegraphics[width=0.9\linewidth]{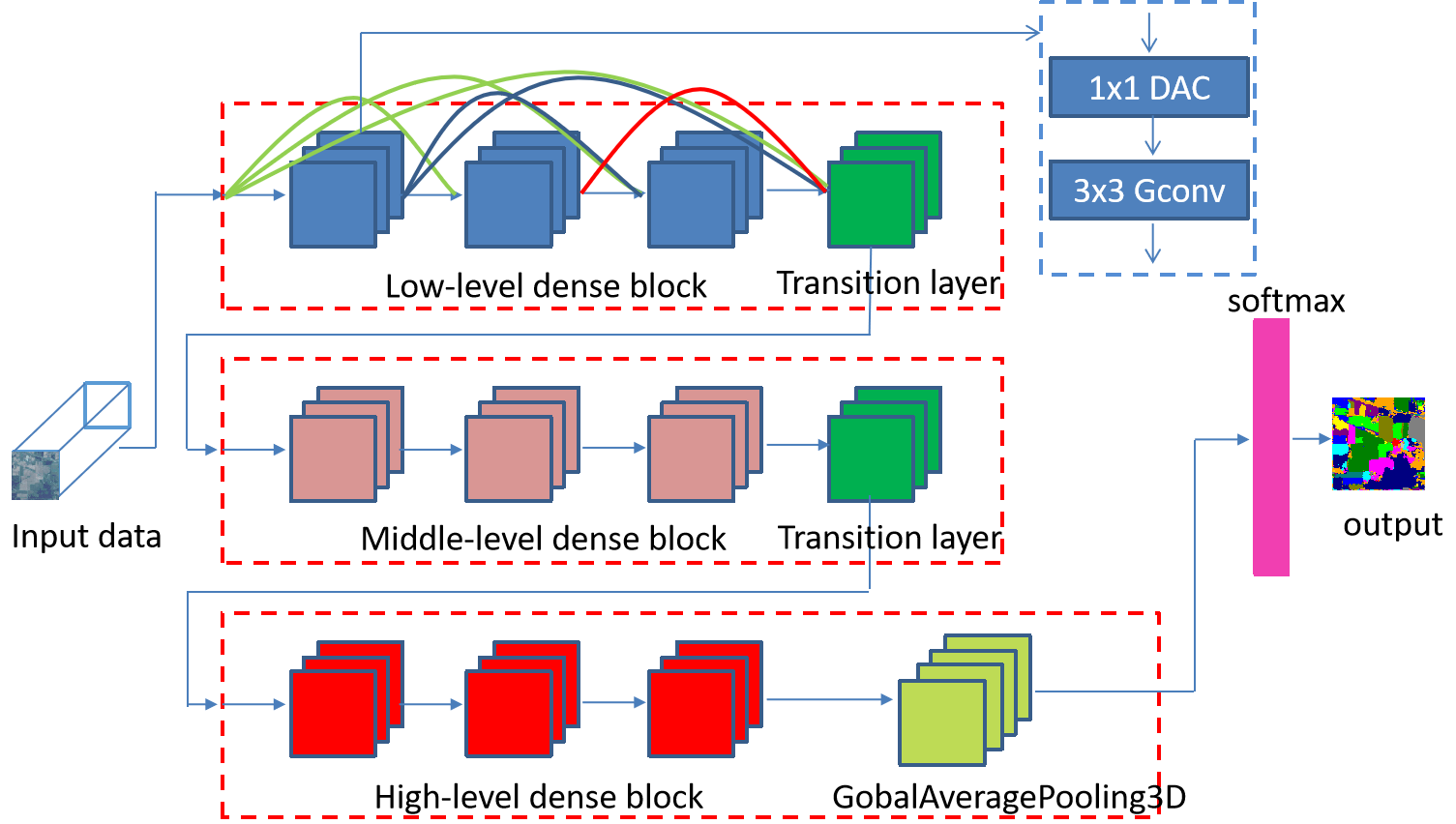}
\caption{Overall architecture of our DACNet, incorporating the modified 3D-DenseNet framework}
\label{fig:dacnet_arch}
\end{figure}

\section{Experiments and Analysis}

To evaluate the performance of DACNet, we conducted experiments on three representative hyperspectral datasets: Indian Pines, Pavia University, and Kennedy Space Center (KSC). The classification metrics include Overall Accuracy (OA), Average Accuracy (AA), and Kappa coefficient.

\subsection{Datasets}
\subsubsection{Indian Pines Dataset}
\begin{figure}[h]
\centering
\includegraphics[width=0.9\linewidth]{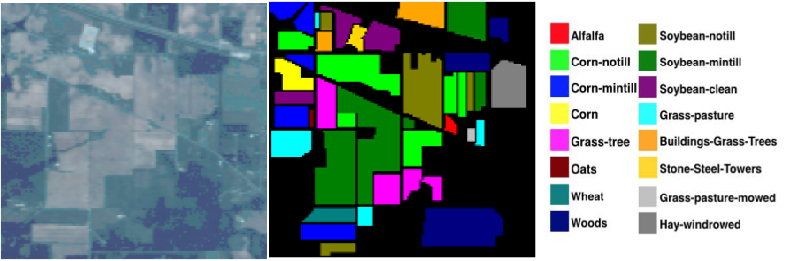}
\caption{False color composite and ground truth labels of Indian Pines dataset}
\label{fig:indian_pines}
\end{figure}

The Indian Pines dataset was collected in June 1992 by the AVIRIS (Airborne Visible/Infrared Imaging Spectrometer) sensor over a pine forest test site in northwestern Indiana, USA. The dataset consists of 145$\times$145 pixel images with a spatial resolution of 20 meters, containing 220 spectral bands covering the wavelength range of 0.4--2.5$\mu$m. In our experiments, we excluded 20 bands affected by water vapor absorption and low signal-to-noise ratio (SNR), utilizing the remaining 200 bands for analysis. The dataset encompasses 16 land cover categories including grasslands, buildings, and various crop types. Figure~\ref{fig:indian_pines} displays the true-color composite image and spatial distribution of ground truth samples, while provides the detailed sample allocation for each land cover category.

\subsubsection{Pavia University Dataset}
\begin{figure}[h]
\centering
\includegraphics[width=0.9\linewidth]{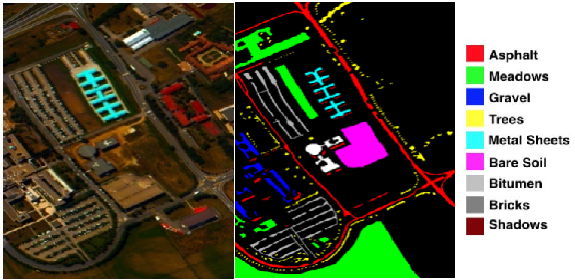}
\caption{False color composite and ground truth labels of Pavia University dataset}
\label{fig:pavia_university}
\end{figure}

The Pavia University dataset was acquired in 2001 by the ROSIS imaging spectrometer over the Pavia region in northern Italy. The dataset contains images of size 610$\times$340 pixels with a spatial resolution of 1.3 meters, comprising 115 spectral bands in the wavelength range of 0.43--0.86$\mu$m. For our experiments, we removed 12 bands containing strong noise and water vapor absorption, retaining 103 bands for analysis. The dataset includes 9 land cover categories such as roads, trees, and roofs. Figure~\ref{fig:pavia_university} shows the spatial distribution of different classes, while provides the detailed sample allocation for each category.

\subsubsection{Kennedy Space Center Dataset}
The KSC dataset was collected on March 23, 1996 by the AVIRIS imaging spectrometer over the Kennedy Space Center in Florida. The AVIRIS sensor captured 224 spectral bands with 10 nm width, centered at wavelengths from 400 to 2500 nm. Acquired from an altitude of approximately 20 km, the dataset has a spatial resolution of 18 meters. After removing bands affected by water absorption and low signal-to-noise ratio (SNR), we used 176 bands for analysis, which represent 13 defined land cover categories.

\subsection{Experimental Analysis}
DACNet was trained for 80 epochs on all three datasets using the Adam optimizer. The experiments were conducted on a platform with four 80GB A100 GPUs. For analysis, we employed the dacnet-base architecture with three stages, where each stage contained 4, 6, and 8 dense blocks respectively. The growth rates were set to 8, 16, and 32, with 4 heads. The 3×3 group convolution used 4 groups, gate\_factor was 0.25, and the compression ratio was 16.

\subsubsection{Data Partitioning Ratio}
For hyperspectral data with limited samples, the training set ratio significantly impacts model performance. To systematically evaluate the sensitivity of data partitioning strategies, we compared the model's generalization performance under different Train/Validation/Test ratios. Experiments show that with limited training samples, a 5:1:4 ratio effectively balances learning capability and evaluation reliability -- this configuration allocates 50\% samples for training, 10\% for validation (enabling early stopping to prevent overfitting), and 40\% for statistically significant testing. DACNet adopted this ratio on Indian Pines, Pavia University and Salinas datasets, with 11×11 neighboring pixel blocks to balance local feature extraction and spatial context integrity.

\begin{table}[h]
\begin{minipage}{\textwidth}
\centering
\makeatletter
\def\@makecaption#1#2{%
    \vskip\abovecaptionskip
    \centering 
    \small #1: #2\par
    \vskip\belowcaptionskip
}
\makeatother
\caption{OA, AA and Kappa metrics for different training set ratios on the Indian Pines dataset}
\begin{adjustbox}{width=0.5\columnwidth}
\begin{tabular}{cccc}
\toprule
Training Ratio & OA & AA & Kappa \\
\midrule
2:1:7 & 92.34 & 91.47 & 91.28 \\
3:1:6 & 97.01 & 96.04 & 96.59 \\
4:1:5 & 98.36 & 96.79 & 98.13 \\
5:1:4 & 95.22 & 96.15 & 94.52 \\
6:1:3 & 99.54 & 99.09 & 99.48 \\
\bottomrule
\end{tabular}
\label{tab:indian_ratios}
\end{adjustbox}
    \end{minipage}

 \vspace*{10pt} 
 
\begin{minipage}{\textwidth}
\centering
\makeatletter
\def\@makecaption#1#2{%
    \vskip\abovecaptionskip
    \centering 
    \small #1: #2\par
    \vskip\belowcaptionskip
}
\makeatother
\caption{OA, AA and Kappa metrics for different training set ratios on the Pavia University dataset}
\begin{tabular}{cccc}
\toprule
Training Ratio & OA & AA & Kappa \\
\midrule
2:1:7 & 99.20 & 98.90 & 98.94 \\
3:1:6 & 99.63 & 99.49 & 99.52 \\
4:1:5 & 99.67 & 99.57 & 99.56 \\
5:1:4 & 99.78 & 99.65 & 99.71 \\
6:1:3 & 99.84 & 99.70 & 99.78 \\
\bottomrule
\end{tabular}
\label{tab:pavia_ratios}
        \end{minipage}

 \vspace*{10pt} 
 
\begin{minipage}{\textwidth}        
\centering
\makeatletter
\def\@makecaption#1#2{%
    \vskip\abovecaptionskip
    \centering 
    \small #1: #2\par
    \vskip\belowcaptionskip
}
\makeatother
\caption{OA, AA and Kappa metrics for different training set ratios on the KSC dataset}
\begin{tabular}{cccc}
\toprule
Training Ratio & OA & AA & Kappa \\
\midrule
2:1:7 & 97.69 & 96.20 & 97.43 \\
3:1:6 & 99.33 & 99.15 & 99.25 \\
4:1:5 & 98.23 & 97.40 & 98.03 \\
5:1:4 & 98.99 & 98.51 & 98.87 \\
6:1:3 & 99.68 & 99.60 & 99.64 \\
\bottomrule
\end{tabular}
\label{tab:ksc_ratios}
        \end{minipage}
\end{table}

\subsubsection{Neighboring Pixel Blocks}
The network pads the input 145×145×103 image (using Indian Pines as example) to 155×155×103, then extracts M×N×L neighboring blocks where M×N is spatial size and L is spectral dimension. Large original images hinder feature extraction, slow processing, and increase memory demands. Thus neighboring block processing is adopted, where block size is a crucial hyperparameter. However, blocks cannot be too small as this limits receptive fields. As shown in Tables \ref{tab:indian_blocks}-\ref{tab:ksc_blocks}, accuracy improves significantly from size 7 to 17 on Indian Pines, with similar trends on Pavia University and KSC datasets. We selected size 17 for all datasets.

\begin{table}[h]
\begin{minipage}{\textwidth}
\centering
\makeatletter
\def\@makecaption#1#2{%
    \vskip\abovecaptionskip
    \centering 
    \small #1: #2\par
    \vskip\belowcaptionskip
}
\makeatother
\caption{OA, AA and Kappa metrics for different block sizes on Indian Pines}
\begin{tabular}{cccc}
\toprule
Block Size (M=N) & OA & AA & Kappa \\
\midrule
7 & 97.98 & 97.07 & 97.70 \\
9 & 98.54 & 97.32 & 98.33 \\
11 & 99.54 & 99.09 & 99.48 \\
13 & 99.80 & 98.93 & 99.78 \\
15 & 99.77 & 99.48 & 99.74 \\
17 & 99.93 & 99.76 & 99.93 \\
\bottomrule
\end{tabular}
\label{tab:indian_blocks}
        \end{minipage}

 \vspace*{10pt} 
 
\begin{minipage}{\textwidth}
\centering
\makeatletter
\def\@makecaption#1#2{%
    \vskip\abovecaptionskip
    \centering 
    \small #1: #2\par
    \vskip\belowcaptionskip
}
\makeatother
\caption{OA, AA and Kappa metrics for different block sizes on Pavia University}
\begin{tabular}{cccc}
\toprule
Block Size (M=N) & OA & AA & Kappa \\
\midrule
7 & 99.89 & 99.82 & 99.86 \\
9 & 99.79 & 99.70 & 99.72 \\
11 & 99.84 & 99.70 & 99.78 \\
13 & 99.95 & 99.94 & 99.93 \\
15 & 99.91 & 99.83 & 99.89 \\
17 & 99.99 & 99.99 & 99.99 \\
\bottomrule
\end{tabular}
\label{tab:pavia_blocks}
        \end{minipage}

 \vspace*{10pt} 
 
\begin{minipage}{\textwidth}
\centering
\makeatletter
\def\@makecaption#1#2{%
    \vskip\abovecaptionskip
    \centering 
    \small #1: #2\par
    \vskip\belowcaptionskip
}
\makeatother
\caption{OA, AA and Kappa metrics for different block sizes on KSC dataset}
\begin{tabular}{cccc}
\toprule
Block Size (M=N) & OA & AA & Kappa \\
\midrule
7 & 98.91 & 98.54 & 98.78 \\
9 & 98.40 & 98.23 & 98.21 \\
11 & 99.68 & 99.60 & 99.64 \\
13 & 99.87 & 99.81 & 99.86 \\
15 & 99.81 & 99.71 & 99.79 \\
17 & 99.94 & 99.90 & 99.93 \\
\bottomrule
\end{tabular}
\label{tab:ksc_blocks}
        \end{minipage}
\end{table}

\subsubsection{Network Parameters}
We categorize DACNet into two variants: base and large. Table~\ref{tab:dacnet_config} shows the OA, AA, and Kappa metrics tested on the Indian Pines dataset. Notably, the large model with higher parameters doesn't show proportional performance gains compared to our efficient base version, demonstrating DACNet's design efficiency.

\begin{table}[h]
\centering
\makeatletter
\def\@makecaption#1#2{%
    \vskip\abovecaptionskip
    \centering 
    \small #1: #2\par
    \vskip\belowcaptionskip
}
\makeatother
\caption{Computational complexity comparison}
\label{tab:param_comparison}
\begin{adjustbox}{width=\textwidth}
\begin{tabular}{lcccccc}
\toprule
Metric & 3D-CNN & 3D-DenseNet & HybridSN & LGCNet & DACNet-base & DACNet-large \\
\midrule
Params (M) & 16.39 & 2.56 & 5.50 & 0.88 & 0.44 & 1.93 \\
GFLOPs & 81.96 & 5.23 & 11.01 & 36.20 & 27.37 & 76.12 \\
\bottomrule
\end{tabular}
\end{adjustbox}
\end{table}

Using the thop library, we measured parameters (Params) and computations (GFLOPs) on Indian Pines (200 spectral dimensions). Table~\ref{tab:param_comparison} shows DACNet-base achieves superior accuracy with minimal parameters.

\begin{table}[h]
\centering
\makeatletter
\def\@makecaption#1#2{%
    \vskip\abovecaptionskip
    \centering 
    \small #1: #2\par
    \vskip\belowcaptionskip
}
\makeatother
\caption{DACNet model configurations and performance metrics}
\label{tab:dacnet_config}
\begin{tabular}{lccccc}  
\toprule
Model & Stages/Blocks & Growth Rate & OA & AA & Kappa \\
\midrule
DACNet-base & 4,6,8 & 8,16,32 & 99.93 & 99.76 & 99.93 \\
DACNet-large & 14,14,14 & 8,16,32 & 99.87 & 98.78 & 99.85 \\
\bottomrule
\end{tabular}
\end{table}

\subsection{Experimental Results and Analysis}
On the Indian Pines dataset, DACNet's input size was set to 17$\times$17$\times$200 (spatial×spatial×spectral). For the Pavia University dataset, LGCNet's input size was 17$\times$17$\times$103, while on the KSC dataset, DACNet's input dimensions were 17$\times$17$\times$176. We evaluated the DACNet-base version and compared it against six state-of-the-art methods: SSRN, 3D-CNN, 3D-SE-DenseNet, Spectralformer, LGCNet, and DGCNet. As shown in Tables~\ref{tab:indian_results}--\ref{tab:pavia_results}, DACNet's three configurations achieved leading accuracy across all evaluation metrics.

\begin{table*}
\begin{minipage}{\textwidth} 
\centering
\makeatletter
\def\@makecaption#1#2{%
    \vskip\abovecaptionskip
    \centering 
    \small #1: #2\par
    \vskip\belowcaptionskip
}
\makeatother
\caption{Classification accuracy (\%) on Indian Pines dataset}
\label{tab:indian_results}
\resizebox{\textwidth}{!}{ 
\begin{tabular}{lccccccc}
\toprule
Class & SSRN & 3D-CNN & 3D-SE-DenseNet & Spectralformer & LGCNet & DGCNet & DACNet \\
\midrule
1 & 100 & 96.88 & 95.87 & 70.52 & 100 & 100 & 100 \\
2 & 99.85 & 98.02 & 98.82 & 81.89 & 99.92 & 99.47 & 100 \\
3 & 99.83 & 97.74 & 99.12 & 91.30 & 99.87 & 99.51 & 100 \\
4 & 100 & 96.89 & 94.83 & 95.53 & 100 & 97.65 & 100 \\
5 & 99.78 & 99.12 & 99.86 & 85.51 & 100 & 100 & 100 \\
6 & 99.81 & 99.41 & 99.33 & 99.32 & 99.56 & 99.88 & 100 \\
7 & 100 & 88.89 & 97.37 & 81.81 & 95.83 & 100 & 100 \\
8 & 100 & 100 & 100 & 75.48 & 100 & 100 & 100 \\
9 & 0 & 100 & 100 & 73.76 & 100 & 100 & 100 \\
10 & 100 & 100 & 99.48 & 98.77 & 99.78 & 98.85 & 100 \\
11 & 99.62 & 99.33 & 98.95 & 93.17 & 99.82 & 99.72 & 99.87 \\
12 & 99.17 & 97.67 & 95.75 & 78.48 & 100 & 99.56 & 100 \\
13 & 100 & 99.64 & 99.28 & 100 & 100 & 100 & 100 \\
14 & 98.87 & 99.65 & 99.55 & 79.49 & 100 & 99.87 & 100 \\
15 & 100 & 96.34 & 98.70 & 100 & 100 & 100 & 100 \\
16 & 98.51 & 97.92 & 96.51 & 100 & 97.73 & 98.30 & 96.43 \\
\midrule
OA & 99.62$\pm$0.00 & 98.23$\pm$0.12 & 98.84$\pm$0.18 & 81.76 & 99.85$\pm$0.04 & 99.58 & 99.93 \\
AA & 93.46$\pm$0.50 & 98.80$\pm$0.11 & 98.42$\pm$0.56 & 87.81 & 99.53$\pm$0.23 & 99.55 & 99.77 \\
K & 99.57$\pm$0.00 & 97.96$\pm$0.53 & 98.60$\pm$0.16 & 79.19 & 99.83$\pm$0.05 & 99.53 & 99.93 \\
\bottomrule
\end{tabular}
}
\end{minipage}

\begin{minipage}{\textwidth} 
\centering
\makeatletter
\def\@makecaption#1#2{%
    \vskip\abovecaptionskip
    \centering 
    \small #1: #2\par
    \vskip\belowcaptionskip
}
\makeatother
\caption{Classification accuracy (\%) on Pavia University dataset}
\label{tab:pavia_results}
\resizebox{\textwidth}{!}{ 
\begin{tabular}{lcccccc}
\toprule
Class & SSRN & 3D-CNN & 3D-SE-DenseNet & Spectralformer & LGCNet & DACNet \\
\midrule
1 & 89.93 & 99.96 & 99.32 & 82.73 & 100 & 100 \\
2 & 86.48 & 99.99 & 99.87 & 94.03 & 100 & 100 \\
3 & 99.95 & 99.64 & 96.76 & 73.66 & 99.88 & 100 \\
4 & 95.78 & 99.83 & 99.23 & 93.75 & 100 & 100 \\
5 & 97.69 & 99.81 & 99.64 & 99.28 & 100 & 100 \\
6 & 95.44 & 99.98 & 99.80 & 90.75 & 100 & 100 \\
7 & 84.40 & 97.97 & 99.47 & 87.56 & 100 & 100 \\
8 & 100 & 99.56 & 99.32 & 95.81 & 100 & 99.91 \\
9 & 90.58$\pm$0.18 & 100 & 100 & 94.21 & 100 & 100 \\
\midrule
OA & 92.99$\pm$0.39 & 99.79$\pm$0.01 & 99.48$\pm$0.02 & 91.07 & 99.99$\pm$0.00 & 99.99 \\
AA & 87.21$\pm$0.25 & 99.75$\pm$0.15 & 99.16$\pm$0.37 & 90.20 & 99.99$\pm$0.01 & 99.99 \\
K & 87.24 & 99.87$\pm$0.27 & 99.31$\pm$0.03 & 88.05 & 99.99$\pm$0.00 & 99.99 \\
\bottomrule
\end{tabular}
}
\end{minipage}
\end{table*}

\begin{figure}[t]
\centering
\includegraphics[width=0.95\linewidth]{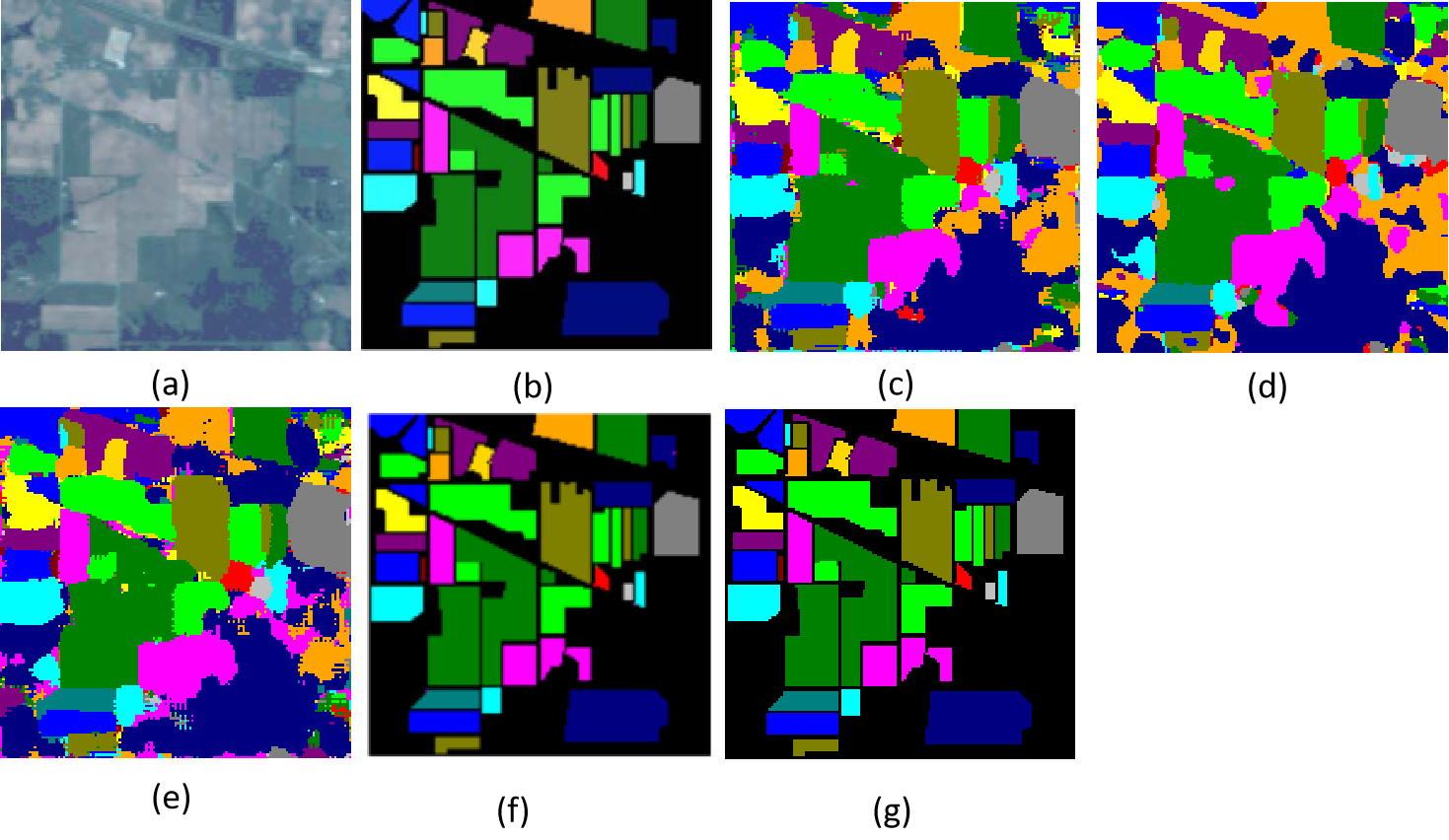}
\caption{Classification results on Indian Pines: (a) False-color image, (b) Ground truth, (c)-(i) Results from comparison methods}
\label{fig:indian_results}
\end{figure}

\section{Conclusion}
This paper introduces Dynamic Attention Convolution (DAC) into 3D convolutional kernels, replacing a single kernel with k parallel convolutional kernels and employing an SE-like attention mechanism to dynamically weight them. This dynamic attention mechanism achieves adaptive feature responses in the spatial dimension of hyperspectral images (HSIs), focusing more effectively on key spatial structures. In the spectral dimension, it enables dynamic discrimination of different bands, mitigating information redundancy and computational complexity caused by high spectral dimensions.Hyperspectral data exhibits sparse characteristics, with uneven sample distribution in the spatial dimension and significant redundancy in the spectral dimension. When extracting features from high-dimensional and redundant HSI data, conventional 3D convolutions also suffer from substantial redundancy among kernels. The DAC module allows 3D-CNNs to prioritize more critical spatial structures and spectral information, ensuring thorough feature extraction while balancing speed and computational efficiency. Moreover, DAC is compatible with existing 3D-CNN architectures, offering both flexibility and high efficiency.

{\small
\bibliographystyle{template}
\bibliography{template}
}

\end{document}